# Mechanical Characterization of Compliant Cellular Robots. Part I: Passive Stiffness

**Gaurav Singh[1]**
Department of Mechanical Engineering and Material Science,
Yale University, New Haven, CT, USA
e-mail: gaurav.singh@yale.edu

**Ahsan Nawroj**
Department of Mechanical Engineering and Material Science,
Yale University, New Haven, CT, USA
e-mail: ahsan.nawroj@gmail.com

**Aaron M. Dollar**
Department of Mechanical Engineering and Material Science,
Yale University, New Haven, CT, USA
e-mail: aaron.dollar@yale.edu

## ABSTRACT

*Modular Active Cell Robots (MACROs) are a design paradigm for modular robotic hardware that uses only two components, namely actuators and passive compliant joints. Under the MACRO approach, a large number of actuators and joints are connected to create mesh-like cellular robotic structures that can be actuated to achieve large deformation and shape-change. In this two-part paper, we study the importance of different possible mesh topologies within the MACRO framework. Regular and semi-regular tilings of the plane are used as the candidate mesh topologies and simulated using Finite Element Analysis (FEA). In Part 1, we use FEA to evaluate their passive stiffness characteristics. Using a strain energy method, the homogenized material properties (Young's modulus, shear modulus, and Poisson's ratio) of the different mesh topologies are computed and compared. The results show that the stiffnesses increase with increasing nodal connectivity and that stretching-dominated topologies have higher stiffness compared to bending-dominated ones. We also investigate the role of relative actuator-node stiffness on the overall mesh characteristics. This analysis shows that the stiffness of stretching-dominated topologies scale directly with their cross-section area whereas bending-dominated ones don't have such a direct relationship.*

---

[1] Corresponding Author

## 1. INTRODUCTION

Modular robots hold promise to enable complex robotic tasks using simple, small, and discrete robotic subunits [1–3]. Such robots are especially promising in resource-constrained environments and in applications where a multitude of tasks has to be accomplished by a small number of robots [2,4,5]. In such cases, modularity and reconfiguration are desirable traits. Modular robots typically consist of multiple identical modules such that each module merges actuation, sensing, and controls in a single self-sufficient unit. This results in these modules being highly complex and difficult to fabricate in large numbers. To address this, we have proposed a new design paradigm for modular robots called Modular Active Cell Robots abbreviated as MACROs [6,7]. MACROs are inspired by the observations in nature of the hierarchical architecture employing simple building blocks resulting in complex biological systems. For example, simple contracting muscle cells are arranged in multiple levels of architectures in skeletal muscles to generate complex motion [8] and in the gastrointestinal tract, simple cells are used in the muscle lining to generate complex anterograde peristalsis [9]. Similarly, the MACRO framework aims to assemble a large number of simple, identical, and easy to fabricate components in a variety of architectures to realize highly articulate, reconfigurable, and modular robots. MACROs can find applications in a variety of fields especially in resource constrained environments such as in outer space or deep sea. In such cases, the MACRO framework has the advantage of using simple and identical components that can be mass-fabricated, thereby simplifying the hardware requirements. Further, the reconfigurability and modularity of MACROs allows the same components to be re-purposed or re-assembled to accomplish a completely different task. MACRO architecture can also help towards creating modular hardware for general-purpose robots that can help with tasks of everyday living. By using the same components and then changing the assembly and topology can lead to a variety of behavior that can include grasping, manipulation, locomotion, etc.

A primary way in which MACROs are simpler than many other modular robot concepts is that they do not have actuators for their own mobility or rearrangement – they are "building blocks" assembled with external actuation (by a human operator, for instance) and accomplish changes to the larger structure by actuating the active cells and thereby causing the surrounding passive compliant elements to strain and cause structural shape change. MACROs are therefore primarily structural robots, designed to provide shape-changing ability to a robotic material. The individual components of the MACRO system are shown in Figure 1.

A physical implementation of the MACRO concept is shown by Nawroj et al. [6,7]. In this implementation, ACs are fabricated using Shape Memory Alloy (SMA) coils and passive bias springs to create a linear-actuating cell that exhibits ~25% strain when actuated (see Figure 1c). Nodes to connect cells to a larger structure are fabricated by moulding Polyurethane in 3D printed moulds (see Figure 1d). Actuation of the mesh can be accomplished in a number of ways, such as controlling individual cells, or in an aggregate way, such as by applying voltage to boundary nodes with internal cells electrically connected. Note that these SMA-based active cells are only one of many ways to create MACROs, and the work described in this paper examines the concept in a way that will generalize to a wide range of designs and implementations.

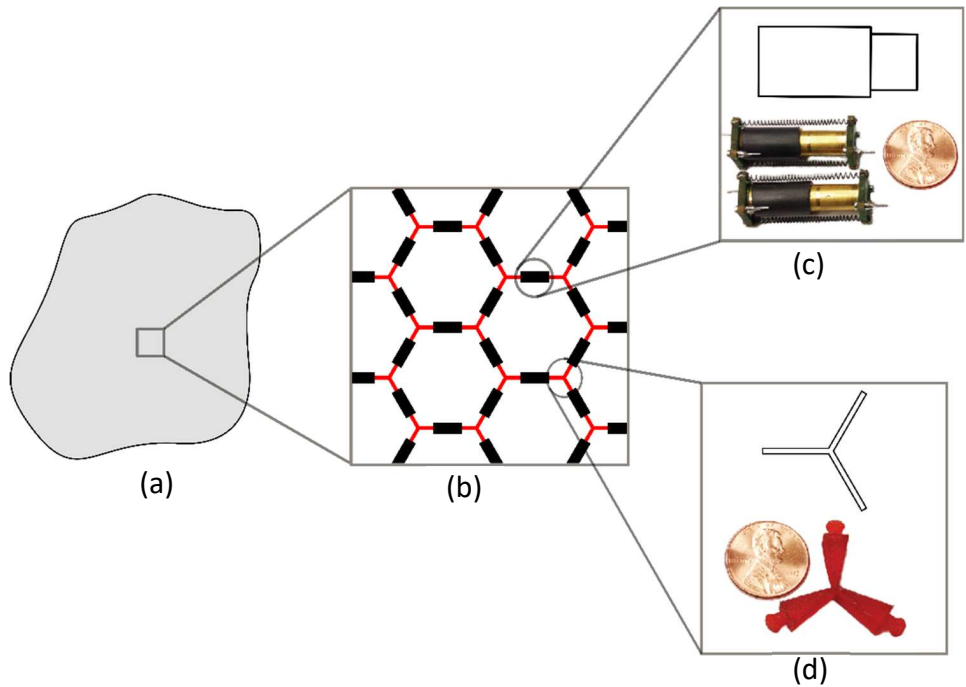

Figure 1. A schematic showing the MACRO framework: (a) Large MACRO robotic material, (b) constitutive MACRO mesh and its components, (c) Active Cells (linear actuators) and (d) passive flexure nodes (joints).

The goal of this paper is to investigate the mechanical properties of different candidate arrangements of active cells into larger structures in order to understand how those choices result in different properties and capabilities of the aggregate structures. We therefore begin by analysing the eleven possible uniform tilings of the plane [10], which are the only possible regular/repeating planar arrangements of units with equal edge length. In Part I of this two-part paper, we look at the passive characteristics of different MACRO meshes, namely their Young's modulus, shear modulus, and Poisson's ratio. In addition, we vary the stiffnesses of actuators and nodes relative to each other to investigate their effects on the properties of overall mesh. Part II of the paper investigates the active characteristics of MACRO meshes or in other words, the strain behavior of a MACRO. In particular, we quantify the overall mesh strain of a large MACRO mesh based on the specific edges of the mesh that are being actuated.

We consider uniform tilings of the plane as candidate topologies for MACRO. Within these tilings, we represent a MACRO using a planar network of edges and vertices (see Figure 1b), with the edges representing simple and identical linear-actuators called Active Cells or ACs [11] as shown in Figure 1c. These ACs are linear actuators that can be either contracting or extending type. The nodes in the MACRO mesh consist of passive elastomeric joints that are compliant and can undergo large bending deformation. These joints can attach to two or more Active Cells depending on the number of arms molded into the node as shown in Figure 1d. Therefore, a large number of ACs are connected to each other through the flexure nodes to create a MACRO mesh that forms the robotic material as shown in Figure 1a-b. A MACRO consists, in addition to the mesh of cells and nodes itself, electrical components to power and actuate the robotic mesh. However, in this two-part paper we are interested in the mechanical characterization of various mesh architectures, so we focus solely on the ACs, the flexure nodes, and their mechanical behavior when assembled in large meshes. Throughout this paper, we use Active Cells (ACs), edges, or actuators interchangeably to refer to the active components in the MACRO mesh, namely the linear actuators. Similarly, we use flexure nodes, joints, vertices or simply nodes to refer to the passive components, namely the passive flexure joints.

To capture the passive behavior of MACROs, we utilize homogenization methods that replace the inhomogeneous material microstructure with a homogeneous medium to evaluate the coefficients of the homogenized elastic tensor $\mathbf{C}^H$ [12,13] and further, evaluate the material properties such as the equivalent Young's modulus, shear modulus, and Poisson's ratio. This approach helps to capture gross behaviors of non-homogenous materials with a repeating microstructure, such as metamaterials, lattice materials or cellular solids [14–17]. Similarly, since the MACRO meshes presented in this paper have a repeating unit cell, we can apply homogenization methods to evaluate their stiffness properties. However, many homogenization methods often require additional postprocessing and are complicated to implement. For our application, since our lattice structures have additional symmetry in their unit cell, we can use an intuitive and relatively easy to implement strain energy method proposed by [18–20]. We briefly summarize this method in Section 2, followed by Section 3 that presents the MACRO mesh topologies considered as the design choices within our framework. Mechanical properties obtained using the strain energy method are presented in Section 4, followed by discussion and conclusions in Sections 5 and 6, respectively.

## 2. STRAIN-ENERGY METHOD

Consider a planar material with a repeating microstructure as shown in Figure 2. If the microstructure of this material has two orthogonal axes of symmetry, then the material is orthotropic. For such a material, we can write the following relation between stresses and strains.

$$\begin{bmatrix} \sigma_{11} \\ \sigma_{22} \\ \sigma_{12} \end{bmatrix} = \begin{bmatrix} C_{1111}^H & C_{1122}^H & 0 \\ C_{2211}^H & C_{2222}^H & 0 \\ 0 & 0 & C_{1212}^H \end{bmatrix} \begin{bmatrix} \varepsilon_{11} \\ \varepsilon_{22} \\ 2\varepsilon_{12} \end{bmatrix} \quad (1)$$

Where the superscript $H$ denotes the homogenized parameters and the subscripts 1 and 2 denote the $X$ and $Y$ axes, respectively. Also, $\sigma$ represents the stresses, $\varepsilon$ represents the strains, and $C^H$ are the coefficients of the stiffness tensor. As shown in Figure 2, since the MACRO structure has two axes of reflective symmetry, only a quarter of this structure needs to be analysed to obtain its material properties. On this symmetric quarter, by applying four different loading conditions as shown in Figure 2 (a-d), we can evaluate the coefficients of the stiffness tensor. These four loading conditions are chosen such that they make all but one of the strains to be equal to zero in Eq. (1), thereby aiding in the evaluation of stiffness coefficients

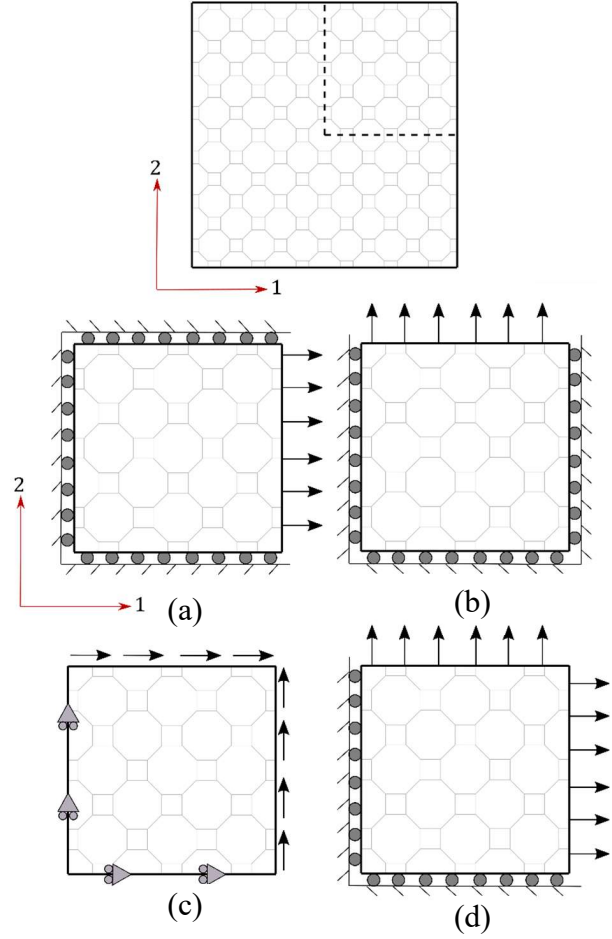

Figure 2. Boundary conditions on a symmetric quarter of the repeating structure to evaluate the coefficients of elastic stiffness tensor.

In the loading condition shown in Figure 2(a), the only non-zero strain is the prescribed strain along the $X$ direction, $\varepsilon_{11}$, while all other strains are zero due to the boundary conditions. Therefore, for this load case, the strain energy is

$$SE^{(a)} = \frac{1}{2}\sigma_{11}\varepsilon_{11} \tag{2}$$

The superscript (a) here denotes the load case shown in Figure 2(a). Further, for this loading condition, we have from Eq. (1)

$$\sigma_{11} = C_{1111}^H \varepsilon_{11} \tag{3}$$

Therefore, from Eq. (2) and (3), we can get

$$C_{1111}^{H} = \frac{2SE^{(a)}}{\left(\varepsilon_{11}\right)^2} \tag{4}$$

Similarly, we can obtain from the remaining three load cases, the following relations between the strain energy, applied strains and the coefficients of the stiffness tensor.

$$C_{2222}^{H} = \frac{2SE^{(b)}}{\left(\varepsilon_{22}\right)^2} \tag{5}$$

$$C_{1212}^{H} = \frac{SE^{(c)}}{2\left(\varepsilon_{12}\right)^2} \tag{6}$$

$$C_{1122}^{H} = \frac{SE^{(d)} - SE^{(a)} - SE^{(b)}}{\varepsilon_{11}\varepsilon_{22}} \tag{7}$$

Note that for the biaxial loading condition shown in Figure 2(d), the prescribed strains $\varepsilon_{11}$ and $\varepsilon_{22}$ have to be equal to the prescribed strains applied in conditions from Figure 2(a) and (b), respectively. This is a necessary condition to obtain the relation shown in Eq. (7).

Next, we can obtain young's modulus, shear modulus, and Poisson's ratio from the coefficients of the stiffness tensor as shown below.

$$E_1 = C_{1111}^{H} - \frac{\left(C_{1122}^{H}\right)^2}{C_{2222}^{H}} \tag{8}$$

$$E_2 = C_{2222}^{H} - \frac{\left(C_{1122}^{H}\right)^2}{C_{1111}^{H}} \tag{9}$$

$$G_{12} = C_{1212}^{H} \tag{10}$$

$$\nu_{12} = \frac{C_{1122}^{H}}{C_{2222}^{H}} \tag{11}$$

$$\nu_{21} = \frac{C_{1122}^{H}}{C_{1111}^{H}} \tag{12}$$

We implement this strain energy method using the commercial FEA package, Abaqus. Timoshenko beam elements (B21 in Abaqus FEA) with rectangular cross-section is used to model both the actuators and nodes. Both the components in the MACRO mesh, i.e., actuators and nodes are assigned the same material properties (Young's modulus, density, and Poisson's ratio), but different cross-section dimensions that capture their different stiffnesses in our model. Since, we are implementing a beam element FE model, this assumption does not lead to any loss in generality. The goal of this study is to investigate the role of different mesh topologies on the resulting MACRO's mechanical properties, irrespective of the type of actuators and compliant joints used. Therefore, we have attempted to keep the model of actuators and nodes in this study as simple as possible. These assumptions allow us to create a computationally inexpensive model of MACROs that allows us to run a large number of simulations of different mesh topologies, loading conditions, and actuation modes.

Figure 3 shows the different steps involved in implementation of the strain energy method in Abaqus. We start with choosing a specific mesh size for the for the mesh topology being analyzed. We use these quantities as inputs for a MATLAB function that outputs the data required to model the MACRO mesh, namely, coordinates of all nodes and a matrix that specifies the nodes that each edge connects to. This data is then imported into Abaqus, where it is used to create the geometry of the part that represents the MACRO mesh. For stiffness simulations, we run four different FE simulations corresponding to the four different loading and boundary conditions shown in Figure 2. Four different Abaqus models are created with the same part geometry and material and section assignments, followed by applying different prescribed displacements and boundary conditions. The four models are solved and the output data is exported to MATLAB for postprocessing. Strain energy, $SE$ and prescribed strains, $\varepsilon$ are imported and then used to evaluate the coefficients of stiffness tensor using Eqs. (4-7). Finally, the equivalent Young's modulus, shear modulus, and Poisson's ratio are calculated using Eqs. (8-12).

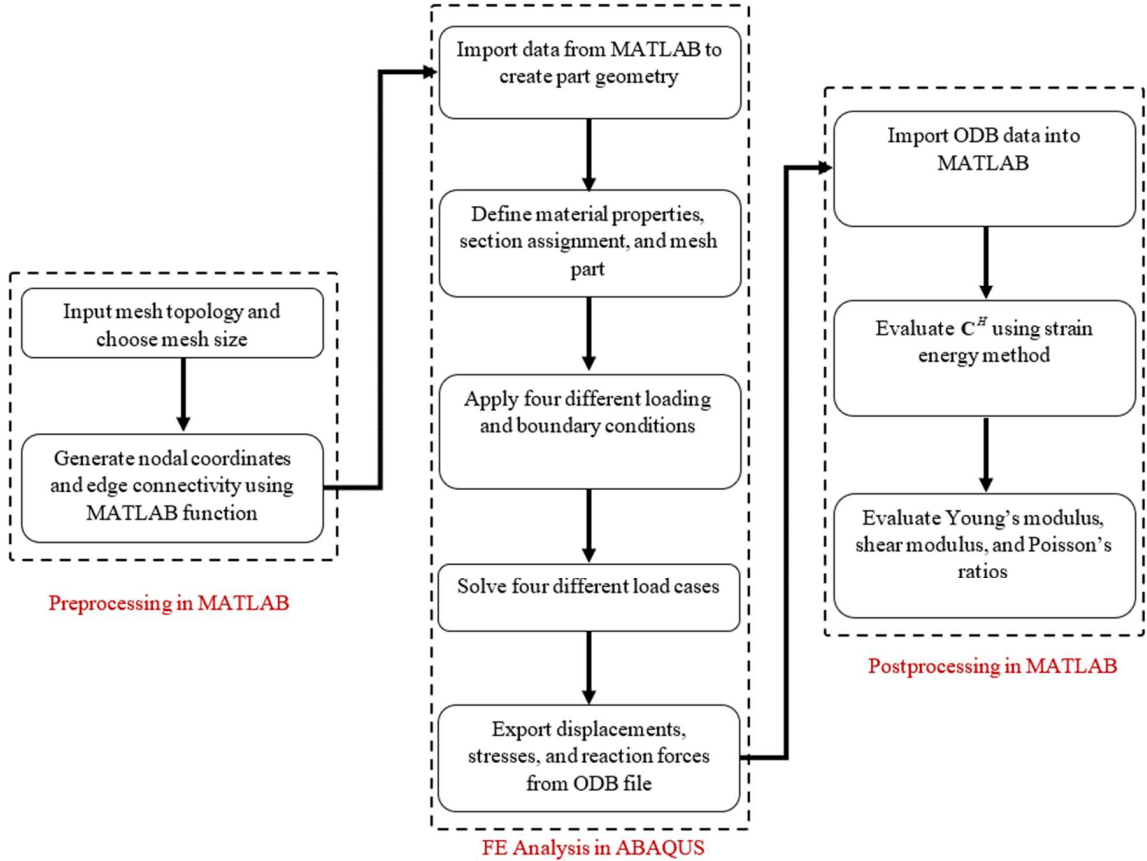

Figure 3. Flow chart showing the different stages involved in the FE analysis to determine the stiffness properties of a MACRO mesh.

In the next section, we describe the design constraints that are imposed on MACROs and the resulting choices of the candidate mesh topologies.

## 3. MESH TOPOLOGIES

Under the MACRO framework, a large number of identical linear actuators are connected to each other using identical compliant joints to form a deformable robotic mesh. We use tilings [10] to denote such a MACRO mesh, where the edges of the tiling mesh denote the actuators and the vertices of the tiling denote the compliant joints connecting the actuators to form the mesh. Due to certain assumptions within the MACRO framework, we have a few design constraints on the possible tiling topologies that can be used for a MACRO. These constraints are listed below. For further details of tilings and associated terminology, refer to Grunbaum and Shephard [10].

- Regularity of enclosed polygons: Since we are constrained to identical actuators, all the edges of the mesh are going to be of the same length at rest. Therefore, each polygon formed by the edges in the mesh are going to be regular i.e., they are both equiangular and equilateral. Regularity of polygons may change after actuation.
- Edge-to-edge tiling: The vertices of the mesh denote the compliant joints connecting the actuators denoted by the edges. Since the actuators can only be connected to other actuators at the compliant joints, all corners and sides of the tiling must coincide with the vertices and edges of the tiling, respectively, thereby implying edge-to-edge tiling.

- Vertex transitivity: All vertices of a MACRO mesh representing the compliant joints are connected to the same number of edges representing the actuators. While not strictly required by the framework, this is a useful generalization for rapid fabrication since only one type of node needs to be fabricated for a given MACRO mesh. A complex mesh that requires non-transitive vertices to account for geometric needs can be considered as a MACRO consisting of two or more MACRO modules, each of which has vertex transitivity.

The mesh topologies that satisfy the above constraints are the uniform tilings of the planar space. There are 11 such tilings and among them there are three regular tilings of the plane implying they use only a single type of regular polygon in their structure. The remaining eight tilings are semi-regular tilings of the plane i.e., they can have more than one type of regular polygon in their structure (although all edges in the mesh are the same length). These 11 regular tilings considered in this paper are shown in Figure 4.

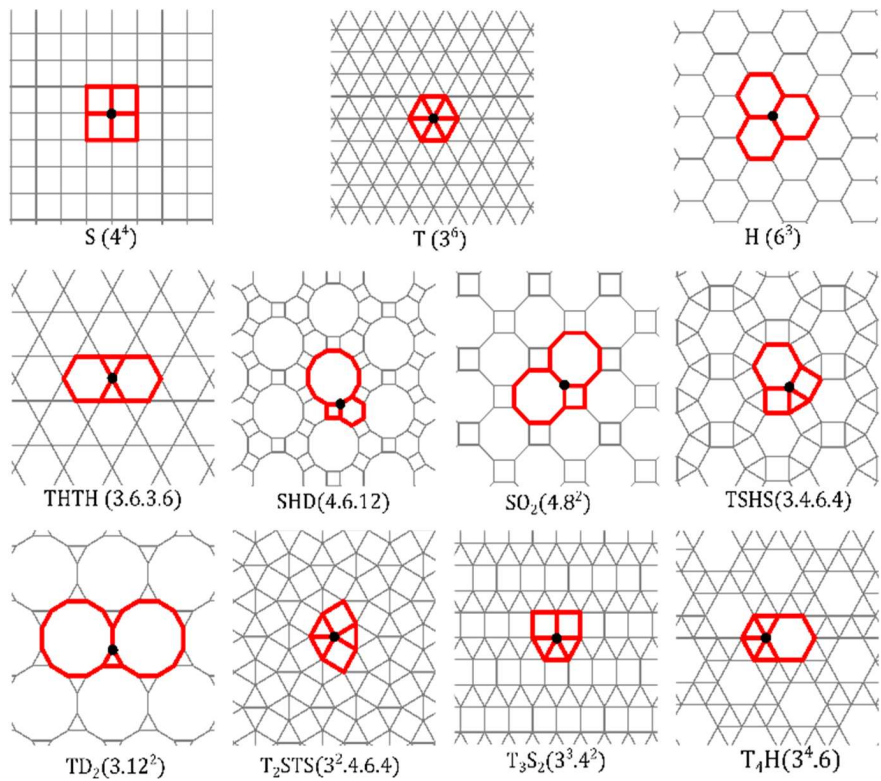

Figure 4. The three uniform regular tilings (top row) and eight uniform semi-regular tilings of the plane [10]. Each of these tiling topologies were used to create mesh topologies for studying the mechanical properties of MACRO meshes.

The vertex transitivity constraint implies that each tiling has only one type of vertex and therefore we denote each tiling based on the vertex's configuration following the convention used in [10]. The naming convention is based on the types of polygons arranged in order around the vertex denoted in black in Figure 4 with the polygons connected to this vertex denoted in red.

A representative volume element (RVE) is the smallest unit of a repetitive structure over which we can make measurements that are representative of the entire structure [21,22]. By choosing an RVE for each mesh such that it has two orthogonal axes of symmetry, we can prove that it is orthotropic and further we can characterize it using the strain energy method. Except for the $T_4H$ mesh, we can choose such an RVE for all the mesh topologies as shown in Figure 5. $T_4H$ type of tiling occurs in two enantiomorphic forms, in other words, it has two mirror images and therefore it does not have reflective symmetry and therefore is not orthotropic. Hence, we have not included $T_4H$ tiling in this stiffness characterization.

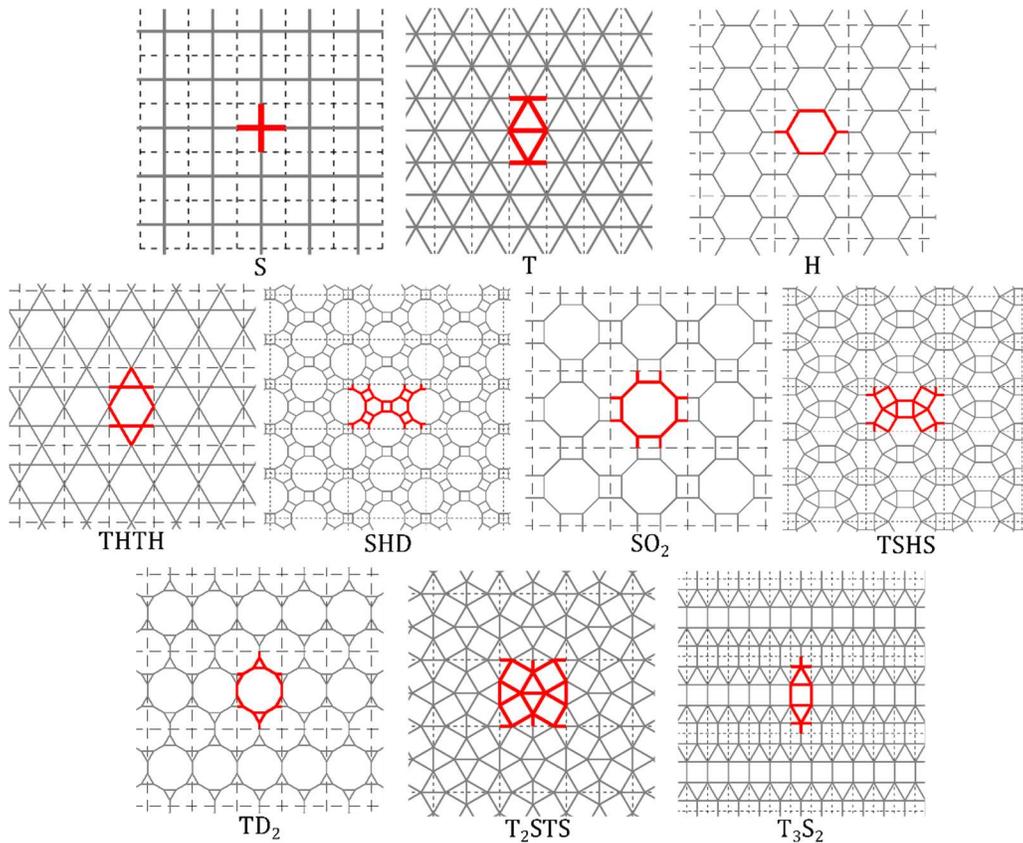

Figure 5. A choice of RVE (shown in red) that is symmetric about two orthogonal axes. These RVEs can also tile the plane by simple translation with no overlap along the aforementioned axes.

## 4. RESULTS

In this section, we present the results obtained from characterizing all the MACRO mesh topologies using the strain energy method. Applying the four different loading conditions as shown in Figure 2 and then using the relations given by Eqs. (4-7), we can obtain the coefficients of the stiffness tensor. Further, we can obtain the homogenized material properties of different topologies, namely Young's modulus, shear modulus, and Poisson's ratio using Eqs. (8-12).

For the FEA simulations, both actuators and nodes are assigned the same Young's modulus of 2000 MPa and Poisson's ratio of 0.3. The difference in the stiffness properties of actuators and nodes is modelled by assigning different cross-section dimensions to these two parts. The in-plane width of actuators is 5 mm while that of nodes is 1 mm. The out of plane width of both components is assigned to be 5 mm. The length of actuators is taken as 25 mm and the length of each arm of a node is taken as 12.5 mm. Therefore, each edge in the MACRO mesh has a length of 50 mm, since it is shared by an actuator and an arm each of the two nodes that this actuator connects.

For each type of mesh topology, we constructed MACRO meshes of four different sizes, 40 MACRO meshes in total. These sizes are chosen to be $750\text{mm} \times 750\text{mm}$, $1000\text{mm} \times 1000\text{mm}$, $1250\text{mm} \times 1250\text{mm}$, and $1500\text{mm} \times 1500\text{mm}$. We applied strains ranging from 1% to 4.5% in the four different loading conditions as mentioned in Section 2 to each of these meshes. We have simulated four meshes of different size to verify that the stiffnesses obtained are independent of the size of the mesh and that there are no boundary effects that influence the obtained modulus values. The average of the Young's modulus in two orthogonal directions and shear modulus obtained for the four different meshes are plotted in Figure 6 for all the MACRO mesh topologies. The scale and limits of the X-axis is the same across all subplots in Figure 6. However, for readability, the Y-axis scale and limits are different for some of the subplots.

For comparison of material properties across all mesh topologies, we have plotted these values at 1% strain in Figure 7. Further, values of Poisson's ratio along two orthogonal directions are plotted in Figure 8.

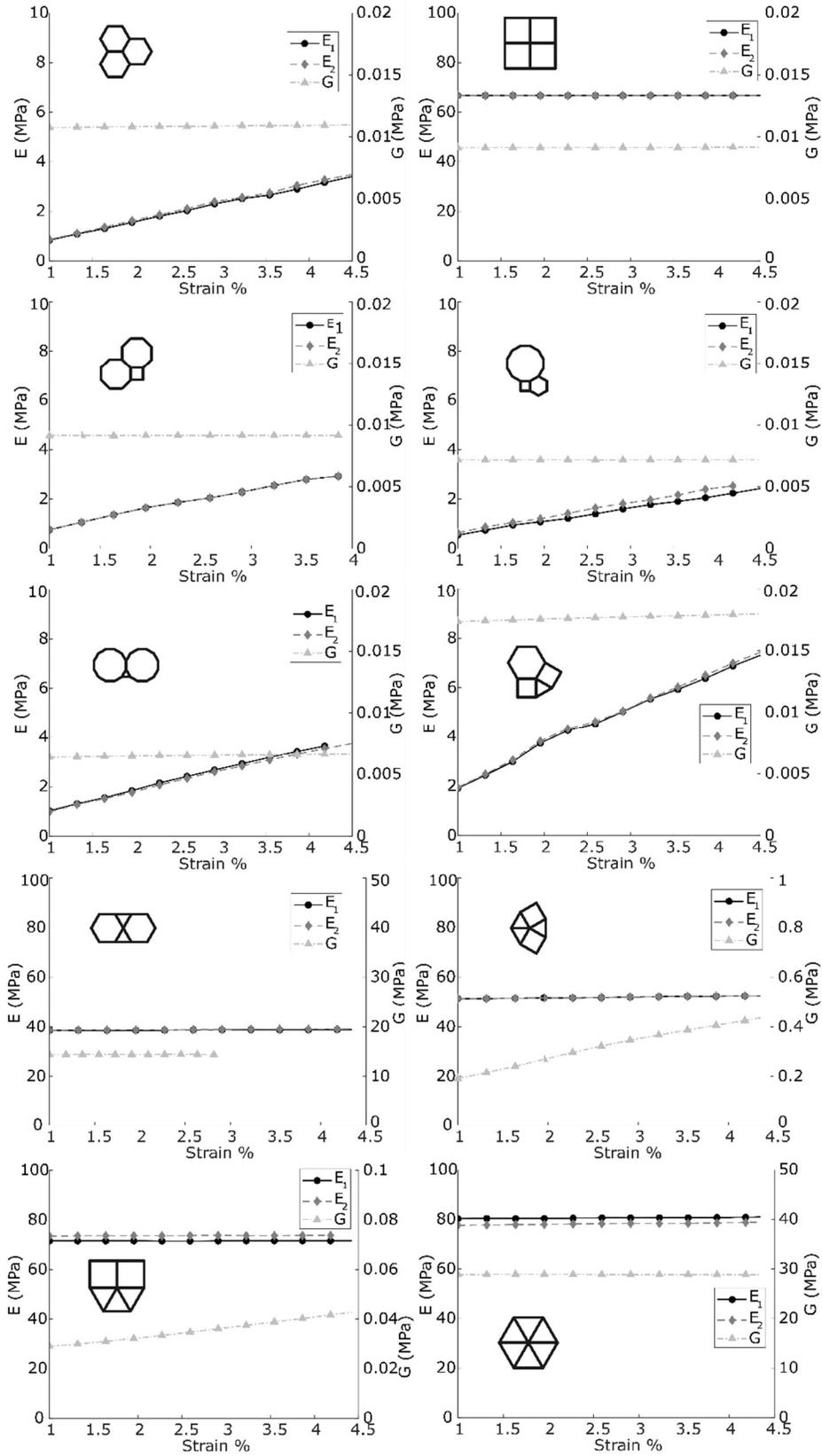

Figure 6. Young's modulus and shear modulus versus applied strain for all 10 mesh topologies with the topology type shown as an inset in the respective plot.

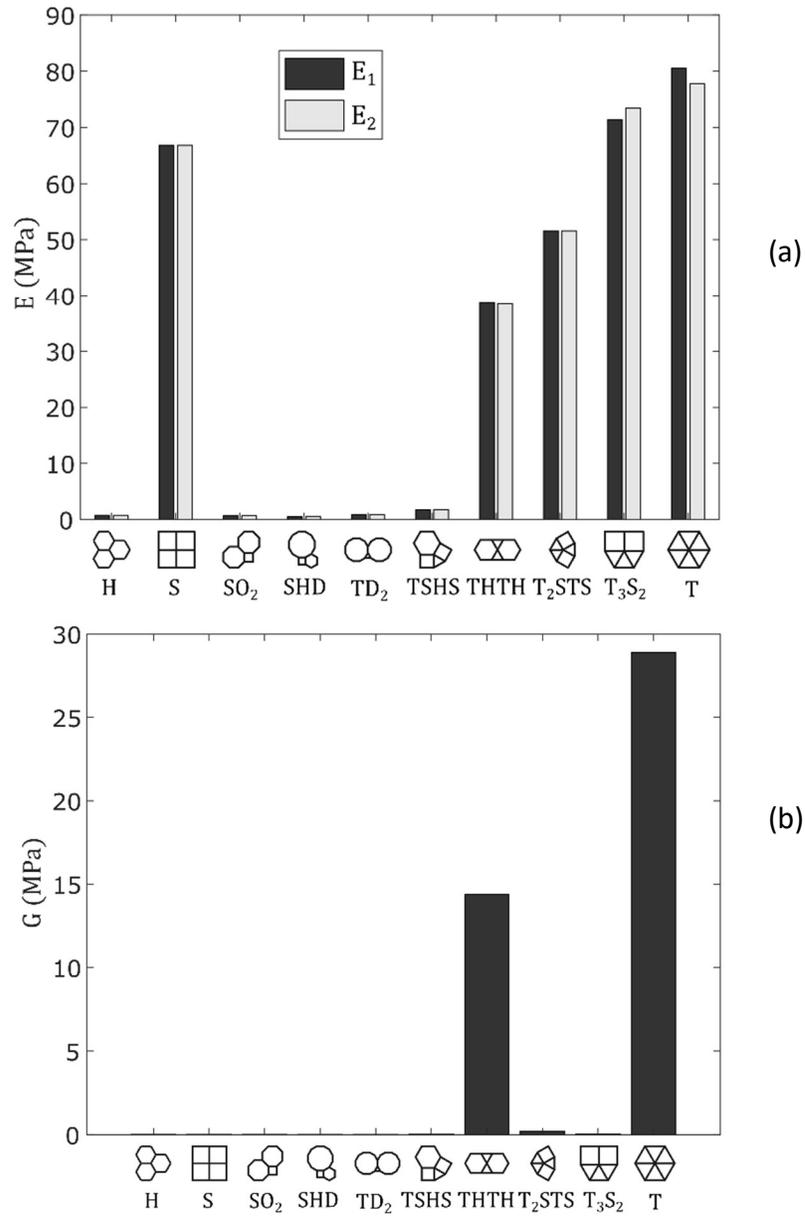

Figure 7. Comparison of material properties at an applied strain of 1%. (a) Young's modulus along $X$ and $Y$ (1 and 2) directions and (b) shear modulus.

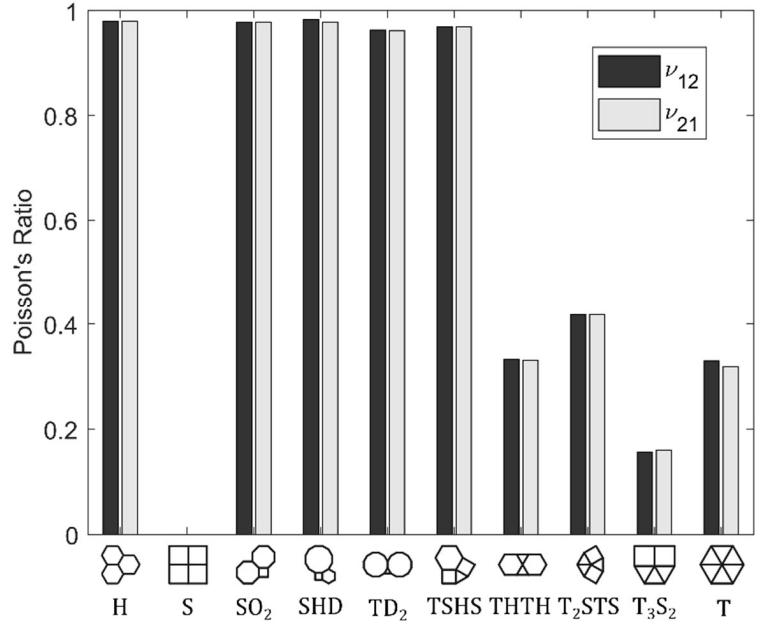

Figure 8. Comparison of Poisson's ratio across all MACRO mesh topologies measured at 1% applied strain.

Next, we simulate three additional MACRO meshes for each type of mesh topology where the relative stiffness of the actuators (i.e., edges) as compared to that of the compliant nodes is varied. We perform these simulations to look at the relationship of overall mesh stiffness properties to that of the stiffnesses of actuators and nodes. In the default case, actuators are stiffer than the nodes and the results for this default case are plotted in Figure 6-Figure 8. The different stiffnesses are modeled by assigning different cross-section dimensions to the actuators as compared to the nodes. In the default case, let the actuator cross-section area be $A_1$ and that of node be $A_2$, where $A_1 = 5A_2$. As mentioned before, we model three additional cases, first case called 'Node Stiff', where the nodes are stiffer than the actuators and in this case cross-section area of nodes is $A_1$ whereas that of actuators is $A_2$. The remaining two cases have identical cross-section and thereby identical stiffnesses for both actuators and nodes, with one case called 'Equal Stiffness (low)' with the assigned cross-section being $A_2$ for both components and the other case called 'Equal Stiffness (high)' with both components assigned area $A_1$.

A heat map of the Young's modulus in the two orthogonal directions for the three additional cases are plotted in Figure 9. Here, the values are divided by the default case where the actuators are stiffer than the nodes. Therefore, a value of one in the heat map implies that there is no change in the overall mesh properties when the properties of the components are varied. Similarly, a heat map of the shear modulus for the three cases are plotted in Figure 10.

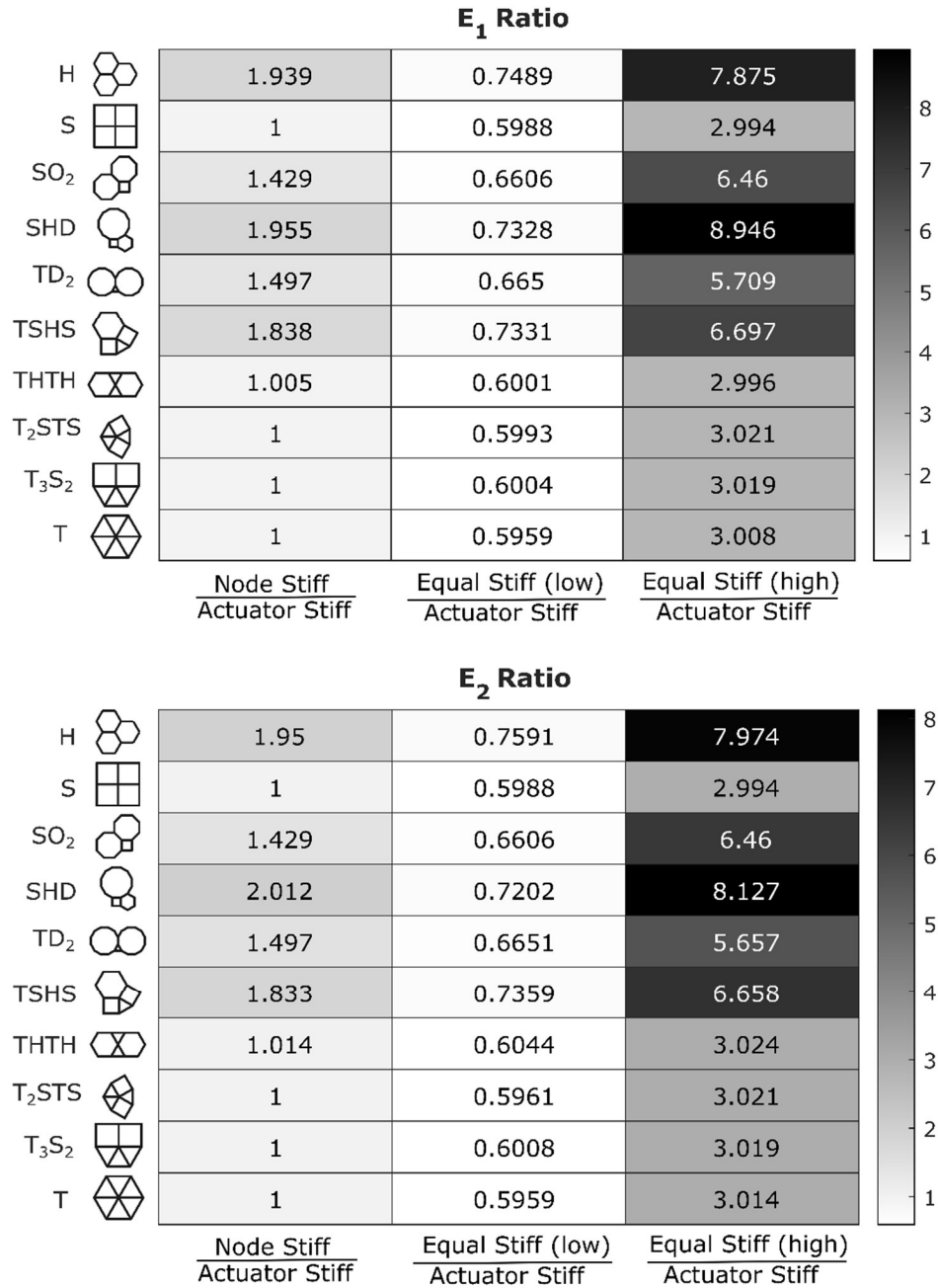

Figure 9. Heat map of Young's modulus at 1% strain for four different conditions of varying the relative stiffness of actuator versus compliant nodes. The values here are divided by the default 'Actuator stiff' configuration.

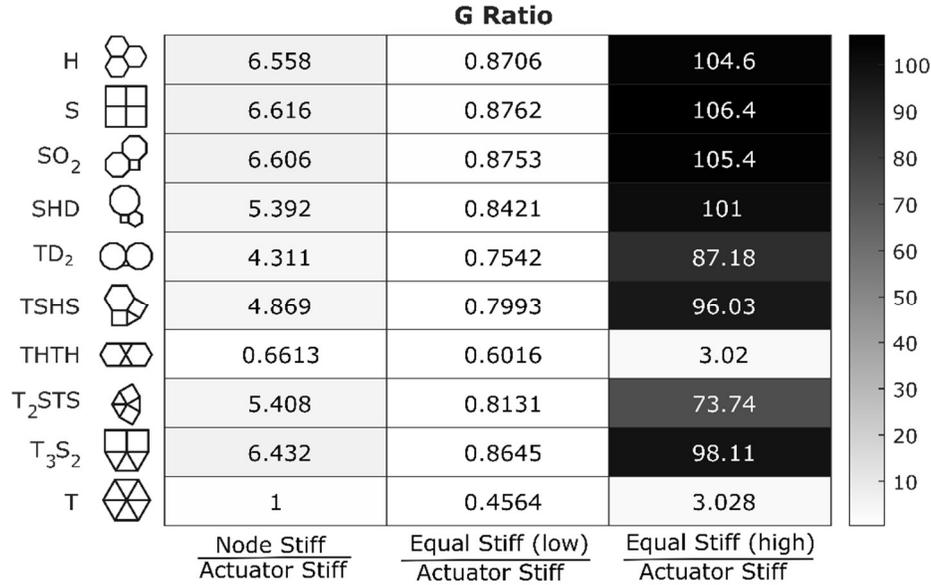

Figure 10. Heat map of shear modulus at 1% strain for four different conditions of varying the relative stiffness of actuator versus compliant nodes. The values here are divided by the default 'Actuator stiff' configuration.

## 5. DISCUSSION

All 10 mesh topologies have values of $E_1$ essentially equivalent to $E_2$ for the range of strains applied, though with some small differences that can be attributed to the numerical errors associated with FEA. This implies that all these topologies have identical stiffnesses in the two orthogonal directions. Hexagon and Triangle honeycombs are isotropic [23] and MACRO meshes are similar in structure to honeycombs except for the difference in cross-section between actuator and nodal sections. Therefore, $E_1$ and $E_2$ being equal indicates isotropic behavior for all topologies, but further validation is required to verify this. Based on the values of Young's modulus obtained, we can rank the topologies as follows: $T > T_3S_2 > S > T_2STS > THTH > TSHS > TD_2 > H > SO_2 > SHD$. Topologies with higher nodal connectivity tend to have higher stiffnesses, which is what we observe here. This rank order mostly follows the order of decreasing nodal connectivity with the exception of $S$ due to the fact that the applied strains in the strain energy method are along the constituent members in the mesh. We can also observe that the first five topologies in this rank order have significantly higher modulus as compared to the remaining topologies. These five topologies undergo stretching-dominated deformation while the remaining undergo bending-dominated deformation. Multiple studies have distinguished lattice materials in these two categories and our findings here are consistent with the literature. Heat maps generated for the stiffness/cross-section area variation studies further support this argument. Stretching-dominated topologies all have the same ratio of $E$, and this ratio is directly proportional to the cross-section area irrespective of whether such area assignment relates to stiffer nodes or stiffer actuators. Bending-dominated topologies, however, do not have a much higher variation in the $E$ ratios. Also, the 'node stiff' case has a higher $E$ value compared to the 'actuator stiff' case for all such topologies. Therefore, having compliant nodes aids deformation only in bending-dominated topologies.

Similarly, we can rank the topologies for shear modulus as follows: $T > THTH > T_2 STS > T_3 S_2 > TSHS > H > S > SO_2 > SHD > TD_2$. Unlike Young's modulus, shear modulus values are significantly higher for only the $T$ and $THTH$ topologies as compared to other topologies. Further, these are the only two topologies that show a linear scaling with cross-section area. Therefore, we can say that only $T$ and $THTH$ show stretching-dominated behavior under shear loading, whereas all other topologies are bending-dominated for shear loading. It should also be pointed out that there is significant increase in $G$ for the 'node stiff' and 'Equal Stiff (high)' cases as compared to the increase in $E$ for these two conditions. This implies that the compliance at the nodes aid in shear deformation of MACRO structures substantially more than they do in the case of deformation along the two orthogonal axes.

The Poisson's ratio plot (see Figure 8) also shows an interesting relation for the bending vs stretching dominated topologies. Bending-dominated topologies all have a Poisson's ratios close to 1, in both directions. In comparison, stretching-dominated topologies have these values in the range of 0.1-0.4.

## 6. CONCLUSIONS AND FUTURE WORK

We modelled and characterized the stiffness properties of different networks of identical linear actuators and compliant nodes in this paper. Under the MACRO framework, we considered uniform tilings of planar space as the design topologies. We evaluated and compared equivalent Young's modulus, shear modulus, and Poisson's ratio for all topologies using a strain energy method. Additionally, we simulated three additional cases by varying the relative stiffness of actuator versus node.

Based on the results, only $T$ and $THTH$ topologies show stretching-dominated behavior in both axial and shear loadings. Therefore, only these two topologies exhibit high stiffness in both axial and shear loading directions. Remaining topologies are bending-dominated in either both or one type of loading, thereby showing a lower stiffness in either one or both the cases. Stretching-dominated behavior shows a direct scaling with the cross-section area, however the location of compliance, whether at the nodes or at the centre of the edges, in such topologies has no effect on its overall stiffness.

In the future, we plan to extend this work to solve the inverse problem of designing a MACRO mesh for a desired stiffness characteristic. Such a problem would involve solving for the mesh topology as well as actuator and nodal properties/geometry for a target equivalent Young's modulus and shear modulus. We further plan to experimentally validate the FE model presented here using different commercially available linear actuators and molded or 3D printed compliant nodes. Finally, extension to spatial lattices of MACRO architecture is another avenue of future work that we plan to undertake.

## FUNDING

This work was supported by the US National Science Foundation under grant 1832795, "EFRI C3 SoRo: Muscle-like Cellular Architectures and Compliant, Distributed Sensing and Control for Soft Robots".

## NOMENCLATURE

| | |
|---|---|
| MACRO | Modular Active Cell Robot |
| AC | Active Cell |
| $\sigma$ | Stress |
| $\varepsilon$ | Strain |
| $\mathbf{C}^H$ | Homogenized elastic tensor |
| $C_{ijkl}^H$ | Coefficient of elastic tensor |
| $SE$ | Strain Energy |
| $E$ | Young's Modulus |
| $G$ | Shear Modulus |
| $\nu$ | Poisson's ratio |
| $A$ | Cross-section area |

# REFERENCES

[1] Chen, I. M., and Yim, M., 2016, "Modular Robots," Springer Handb. Robot., pp. 531–542.

[2] Romanishin, J. W., Gilpin, K., and Rus, D., 2013, "M-Blocks: Momentum-Driven, Magnetic Modular Robots," IEEE Int. Conf. Intell. Robot. Syst., pp. 4288–4295.

[3] White, P., Zykov, V., Bongard, J., and Lipson, H., 2005, "Three Dimensional Stochastic Reconfiguration of Modular Robots," Robot. Sci. Syst., **1**, pp. 161–168.

[4] Zykov, V., Chan, A., and Lipson, H., 2007, "Molecubes: An Open-Source Modular Robotics Kit," IROS-2007 Self-Reconfigurable Robot. Work., pp. 3–6.

[5] Murata, S., Yoshida, E., Kamimura, A., Kurokawa, H., Tomita, K., and Kokaji, S., 2002, "M-TRAN : Self-Reconfigurable Modular," **7**(4), pp. 431–441.

[6] Nawroj, A. I., and Dollar, A. M., 2017, "Shape Control of Compliant, Articulated Meshes: Towards Modular Active-Cell Robots (MACROs)," IEEE Robot. Autom. Lett., **2**(4), pp. 1878–1884.

[7] Nawroj, A. I., Swensen, J. P., and Dollar, A. M., 2017, "Toward Modular Active-Cell Robots (MACROs): SMA Cell Design and Modeling of Compliant, Articulated Meshes," IEEE Trans. Robot., **33**(4), pp. 796–806.

[8] Lieber, R. L., and Fridén, J., 2000, "Functional and Clinical Significance," Muscle Nerve, **23**(November), pp. 1647–1666.

[9] Hirst, G. D. S., 1979, "Mechanisms of Peristalsis," Br. Med. Bull., **35**(3), pp. 263–268.

[10] Grunbaum, B., and Shephard, G. C., 2019, "Tilings by Regular Polygons," Math. Mag., **50**(5), pp. 227–247.

[11] Nawroj, A. I., Swensen, J. P., and Dollar, A. M., 2015, "Design of Mesoscale Active Cells for Networked, Compliant Robotic Structures," *2015 IEEE/RSJ International Conference on Intelligent Robots and Systems (IROS)*, IEEE, pp. 3284–3289.

[12] Nguyen, V. D., and Noels, L., 2014, "Computational Homogenization of Cellular Materials," Int. J. Solids Struct., **51**(11–12), pp. 2183–2203.

[13] Hassani, B., and Hinton, E., 1998, "A Review of Homogenization and Topology Optimization I - Homogenization Theory for Media with Periodic Structure," Comput. Struct., **69**(6), pp. 707–717.

[14] Wagner, M. A., Lumpe, T. S., Chen, T., and Shea, K., 2019, "Programmable, Active Lattice Structures: Unifying Stretch-Dominated and Bending-Dominated Topologies," Extrem. Mech. Lett., **29**, p. 100461.

[15] Bertoldi, K., Vitelli, V., Christensen, J., and Van Hecke, M., 2017, "Flexible Mechanical Metamaterials," Nat. Rev. Mater., **2**.

[16] Barchiesi, E., Spagnuolo, M., and Placidi, L., 2019, "Mechanical Metamaterials: A State of the Art," Math. Mech. Solids, **24**(1), pp. 212–234.

[17] Haghpanah, B., Ebrahimi, H., Mousanezhad, D., Hopkins, J., and Vaziri, A., 2016, "Programmable Elastic Metamaterials," Adv. Eng. Mater., **18**(4), pp. 643–649.

[18] Zhang, W., Dai, G., Wang, F., Sun, S., and Bassir, H., 2007, "Using Strain Energy-Based Prediction of Effective Elastic Properties in Topology Optimization of Material Microstructures," Acta Mech. Sin., **23**(1), pp. 77–89.

[19] Patiballa, S. K., and Krishnan, G., 2018, "Qualitative Analysis and Conceptual

Design of Planar Metamaterials With Negative Poisson's Ratio," J. Mech. Robot., **10**(2), p. 021006.

[20]    Mehta, V., Frecker, M., and Lesieutre, G. A., 2009, "Topology Optimization of Contact-Aided Compliant Cellular Mechanisms," *Proceedings of the ASME 2009 Conference on Smart Materials, Adaptive Structures and Intelligent Systems, SMASIS*, Oxnard, California.

[21]    Alsayednoor, J., Harrison, P., and Guo, Z., 2013, "Large Strain Compressive Response of 2-D Periodic Representative Volume Element for Random Foam Microstructures," Mech. Mater., **66**, pp. 7–20.

[22]    Sun, C.-T., and Vaidya, R. S., 1996, "Prediction of Composite Properties from a Representative Volume Element," Compos. Sci. Technol., **56**(2), pp. 171–179.

[23]    Gibson, L. J., and Ashby, M. F., 1997, *Cellular Solids: Structure and Properties*, Cambridge University Press, Cambridge.

**Figure Captions List**

Figure 1. A schematic showing the MACRO framework: (a) Large MACRO robotic material, (b) constitutive MACRO mesh and its components, (c) Active Cells (linear actuators) and (d) passive flexure nodes (joints).

Figure 2. Boundary conditions on a symmetric quarter of the repeating structure to evaluate the coefficients of elastic stiffness tensor.

Figure 3. Flow chart showing the different stages involved in the FE analysis to determine the stiffness properties of a MACRO mesh.

Figure 4. The three uniform regular tilings (top row) and eight uniform semi-regular tilings of the plane [10]. Each of these tiling topologies were used to create mesh topologies for studying the mechanical properties of MACRO meshes.

Figure 5. A choice of RVE (shown in red) that is symmetric about two orthogonal axes. These RVEs can also tile the plane by simple translation with no overlap along the aforementioned axes.

Figure 6. Young's modulus and shear modulus versus applied strain for all 10 mesh topologies with the topology type shown as an inset in the respective plot.

Figure 7. Comparison of material properties at an applied strain of 1%. (a) Young's modulus along $X$ and $Y$ (1 and 2) directions and (b) shear modulus.

Figure 8. Comparison of Poisson's ratio across all MACRO mesh topologies measured at 1% applied strain.

Figure 9. Heat map of Young's modulus at 1% strain for four different conditions of varying the relative stiffness of actuator versus compliant nodes. The values here are divided by the default 'Actuator stiff' configuration.

Figure 10. Heat map of shear modulus at 1% strain for four different conditions of varying the relative stiffness of actuator versus compliant nodes. The values here are divided by the default 'Actuator stiff' configuration.